\documentclass[final]{patem}

\usepackage{times}
\usepackage{epsfig}
\usepackage{graphicx}
\usepackage{amsmath}
\usepackage{amssymb}

\usepackage[pagebackref=true,breaklinks=true,colorlinks,bookmarks=false]{hyperref}

\begin{document}

%%%%%%%%% TITLE
\title{ Hyperdimensional computing as a framework for \\ 
systematic aggregation of image descriptors}

\author{Peer Neubert\\
TU Chemnitz\\
Germany\\
{\tt\small peer.neubert@etit.tu-chemnitz.de}
\and
Stefan Schubert\\
TU Chemnitz\\
Germany\\
{\tt\small stefan.schubert@etit.tu-chemnitz.de}
}

\maketitle

%%%%%%%%% ABSTRACT
\begin{abstract}

Image and video descriptors are an omnipresent tool in computer vision and its application fields like mobile robotics.
Many hand-crafted and in particular learned image descriptors are numerical vectors with a potentially (very) large number of dimensions.
Practical considerations like memory consumption or time for comparisons call for the creation of compact representations.
In this paper, we use hyperdimensional computing (HDC) as an approach to systematically combine information from a set of vectors in a single vector of the same dimensionality. 
HDC is a known technique to perform symbolic processing with distributed representation in numerical vectors with thousands of dimensions.
We present a HDC implementation that is suitable for processing the output of existing and future (deep-learning based) image descriptors.
We discuss how this can be used as a framework to process descriptors together with additional knowledge by simple and fast vector operations. 
A concrete outcome is a novel HDC-based approach to aggregate a set of local image descriptors together with their image positions in a single holistic descriptor.
The comparison to available holistic descriptors and aggregation methods on a series of standard mobile robotics place recognition experiments shows a 20\% improvement in average performance compared to runner-up and 3.6x better worst-case performance.

\end{abstract}

%%%%%%%%% BODY TEXT
\section{Introduction}

Image descriptors are very useful tools for recognition tasks in computer vision.
Many hand crafted and in particular deep learning based descriptors are numerical vectors with a potentially large number of dimensions, e.g. NetVLAD \cite{netvlad} uses 4,096-D vectors (after PCA), DELF \cite{delf} uses 1,024-D vectors (before PCA). 
Approaches like BoW \cite{SivicZ06}, VLAD \cite{vlad}, or ASMK \cite{ToliasAJ16} aggregate the information from multiple vectors in a single holistic vector representation to reduce memory consumption and computational efforts during comparison.
For example, deciding whether two images show the same place based on a set of local landmarks from each image, can then be done by a single distance measure between the two aggregated vectors.
Although these techniques are able to combine large numbers of descriptors in a compact vector, for certain tasks, like place recognition, it is beneficial to encode additional information in the final vector representation, e.g., information about the image locations of aggregated vectors.

The central idea of this paper is to use binding and bundling of vectors as a flexible framework to combine image descriptors and additional information.
The underlying technique of binding and bundling vectors is taken from a field known as hyperdimensional computing (HDC) or vector symbolic architectures (VSA).
This is an established class of approaches to solve \textit{symbolic} computational problems using mathematical operations on large numerical vectors with thousands of dimensions \cite{Kanerva09,Neubert19}. 
The bundling operator $\oplus$  superposes information of a variable number of vectors in a single vector; we can think of it as some form of averaging.
The binding operator $\otimes$ can, for example, express role-filler or variable-value pairs as required in symbolic processing. 
An important property is that the output of the operations are vectors from the same vector space. 
This allows to chain HDC operations and enables versatile encoding of structured data from a set of d-dimensional vectors in a single d-dimensional vector.

We will present a HDC implementation that allows the processing of existing and future (deep learning based) image descriptors in Sec.~\ref{sec:approach}.
This section will also describe how HDC can be used as a framework to aggregate holistic or local image descriptors and to combine them with additional information.
A concrete outcome is a novel approach to create a holistic image descriptor from a set of local descriptors with position information in Sec.~\ref{sec:approach_local}.
For example, we can create a holistic descriptor from three local descriptors $L_1, L_2, L_3$ with poses $P_1, P_2, P_2$ as simple as $(L_1 \otimes P_1) \oplus (L_2 \otimes P_2) \oplus (L_3 \otimes P_3)$.
The poses serve as ``roles`` that are associated with landmarks as ``fillers``.
When comparing two such holistic descriptors (e.g. by using simple cosine similarity), the similarity of the roles decides to what extend the similarities of the associated local descriptors are incorporated in the overall similarity. 
Prerequisite are appropriate preprocessing of the descriptors as well as a suitable encoding of image positions in the same vector space as the descriptors, both will be presented in Sec.~\ref{sec:our_vsa}.
The experiments in Sec.~\ref{sec:experiments} will evaluate properties in a series of mobile robotics place recognition experiments.

%%%%%%%%% BODY TEXT
\section{Related Work}

\subsection{Descriptors for place recognition}

Visual place recognition \cite{Lowry16} is a basic problem in mobile robotics, e.g., for loop closure detection in SLAM or candidate selection for visual localization \cite{Sattler2018}. 
In contrast to \mbox{6-DOF} pose estimation that often uses local features (e.g. keypoints \cite{Lowe04,Detone18,delf,d2net}), place recognition typically builds upon holistic image descriptors that compute a single descriptor vector for a whole image \cite{netvlad,Torii15,suenderhauf15,seqslam,NeubertSP19}. 
Important reasons are the memory consumption and the required time for exhaustively comparing local features.

The existing and steadily increasing (e.g., \cite{Revaud19,Zixin20,RFNET}) variety of local feature extractors can also be the basis for holistic image descriptors.
Existing approaches include BoW \cite{SivicZ06,CumminsN11}, Fisher vectors \cite{PerronninD07}, and VLAD \cite{vlad,ArandjelovicZ13}. 
Aggregated selective match kernels \cite{ToliasAJ16} aim at unifying aggregation-based techniques with matching-based approaches like Hamming Embedding \cite{JegouDS10}. 
VLAD in combination with soft-assignment is fully differentiable and seamlessly integrates in deep learning approaches, e.g. NetVLAD \cite{netvlad}. 
Other deep learning variants of local feature aggregation for image matching include sum-pooling \cite{ToliasJC20}, max-pooling \cite{HusainB19}, and mean-pooling \cite{CaoAS20}. 
The latter also outputs global and local descriptors.

Besides their descriptors, the \textit{location} of local features can provide important information, e.g. for geometric verification \cite{SattlerLK12}.
Regarding holistic descriptors, BoW can integrate spatial information via voting \cite{ShenLBAW12}.
Pyramid match kernels \cite{pyramidmatchkernel} can evaluate matchings at multiple resolutions. 
Based on this, spatial pyramid matching \cite{spatialpyramid} can approximate global geometric correspondence between sets of local features. 
Multi-VLAD \cite{ArandjelovicZ13} extends this idea to VLAD, Pyramid-Enhanced NetVLAD \cite{SPEVLAD} extends it to deep learning.
The typical usage of flattened AlexNet-conv3 \cite{Krizhevsky12} descriptors (or similar) as in \cite{suenderhauf15}\cite{SchubertNP20} is an implicit encoding of local landmarks (one landmark per feature map vector) together with their image location (encoded by the position in the concatenated output vector). Similar encodings can be applied to other local features. 

To reduce memory consumption and runtime for comparisons, descriptors are often combined with dimensionality reduction approaches like PCA \cite{delf} or Gaussian random projections \cite{suenderhauf15}, or compression techniques like product quantization \cite{JegouDS11}. 
Approximate nearest neighbor and inverted indexes play an important role for large-scale image matching \cite{flann,Li2020,SivicZ06}.

\subsection{Hyperdimensional Computing}

Hyperdimensional computing (HDC) is also known as Vector Symbolic Architectures (VSA) or computing with large random vectors.
It is an established class of approaches to solve \textit{symbolic} computational problems using mathematical operations on large numerical vectors with thousands of dimensions \cite{Kanerva09,Plate94Phd,gayler03,Schlegel2020}. 
Using embeddings in high-dimensional vector spaces to deal with ambiguities is well established in natural language processing \cite{Camacho-Collados18}. 
HDC makes use of additional operations on high-dimensional vectors. 
So far, HDC has been applied in various fields including addressing catastrophic forgetting in deep neural networks \cite{CheungTCAO19}, medical diagnosis \cite{Widdows15}, robotics \cite{Neubert19}, fault detection \cite{Kleyko15a}, analogy mapping \cite{Rachkovskij12}, reinforcement learning \cite{Kleyko15}, long-short term memory \cite{Danihelka16}, text classification \cite{Kleyko18}, and synthesis of finite state automata \cite{Osipov17}.
They have been used in combination with deep-learned descriptors before, e.g. for sequence encoding \cite{Neubert19}.
Another related HDC work are spatial semantic pointers \cite{KomerSVE19}, a variant of the Semantic Pointer Architecture \cite{Eliasmith07}, that processes vector encodings of symbols with positions in images using a complex vector space and fractional binding.

%%%%%%%%% BODY TEXT
\section{Algorithmic Approach}
\label{sec:approach}

We will first describe the basic elements of the proposed HDC framework in Sec.~\ref{sec:our_vsa} followed by examples, how these elements can be used to approach image retrieval tasks with different types of available information in Sec.~\ref{sec:feat_aggr}.
% ============================================================================
% ============================================================================
\subsection{HDC architecture}
\label{sec:our_vsa}

In simple words, we use elementwise addition and elementwise multiplication of 4,096 dimensional real vectors of small magnitude (typical vector entries are in range $[-1,1]$) to systematically encode information. The vectors are either systematic encodings of systematic information (e.g., images or distances) or random vectors for discrete symbols (e.g., a descriptor type). 
The way how we process vectors is borrowed from the HDC literature. We will use the HDC terminology of binding and bundling operators for the multiplication and addition operators. The 
main reason is that there are other HDC implementations available that implement binding and bundling differently, and which could be used to replace our particular HDC architecture for the later presented aggregation approaches in Sec.~\ref{sec:feat_aggr}. 
Our implementation is similar to the Multiply-Add-Permute (MAP) architecture by Gaylor \cite{Gayler98,gayler03}. 
However, we change the vector space for combination with image descriptors which has some ramifications for the operators as well. 

\begin{figure}[t]
 \includegraphics[width=0.324\linewidth]{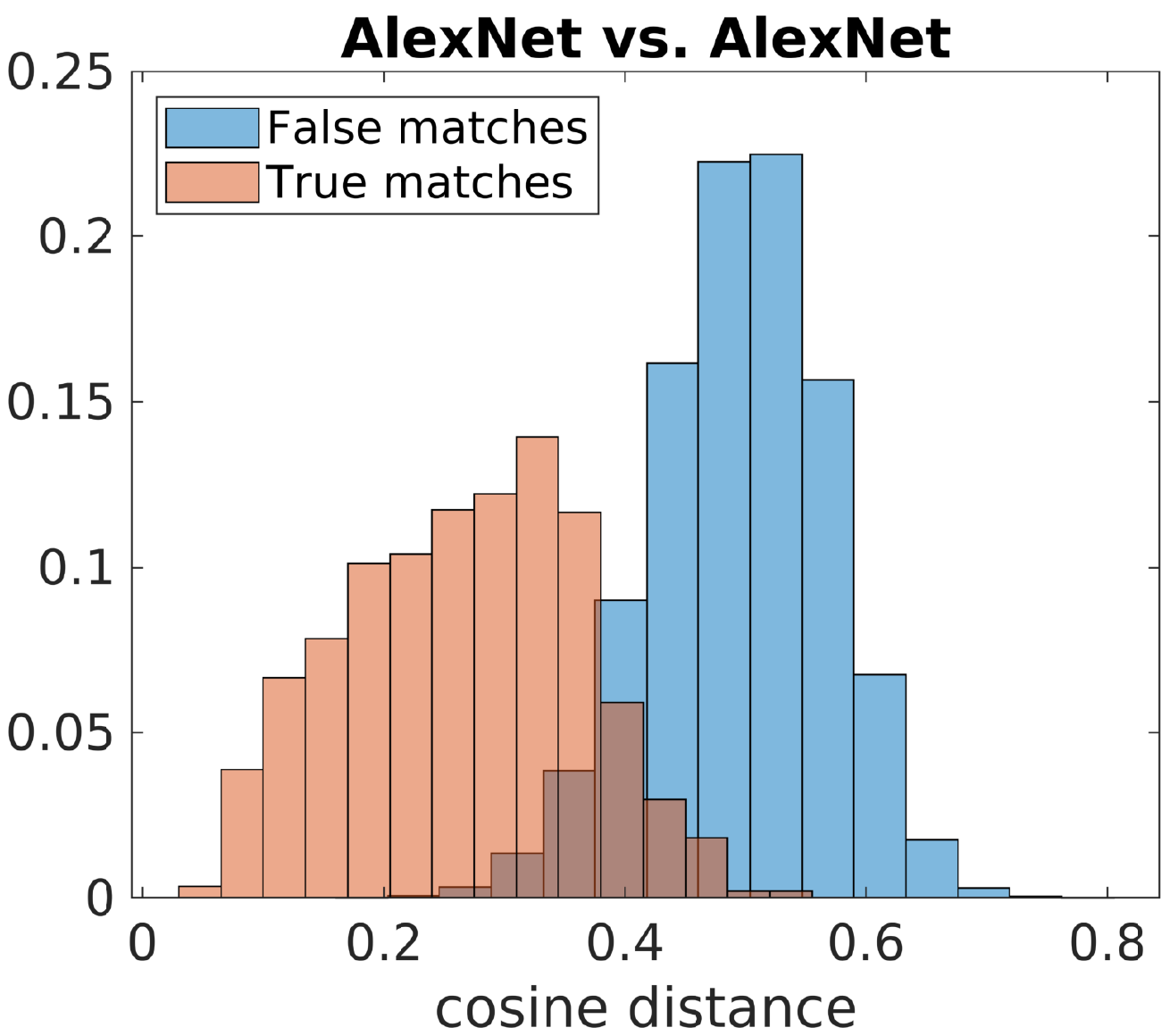}
 \includegraphics[width=0.324\linewidth]{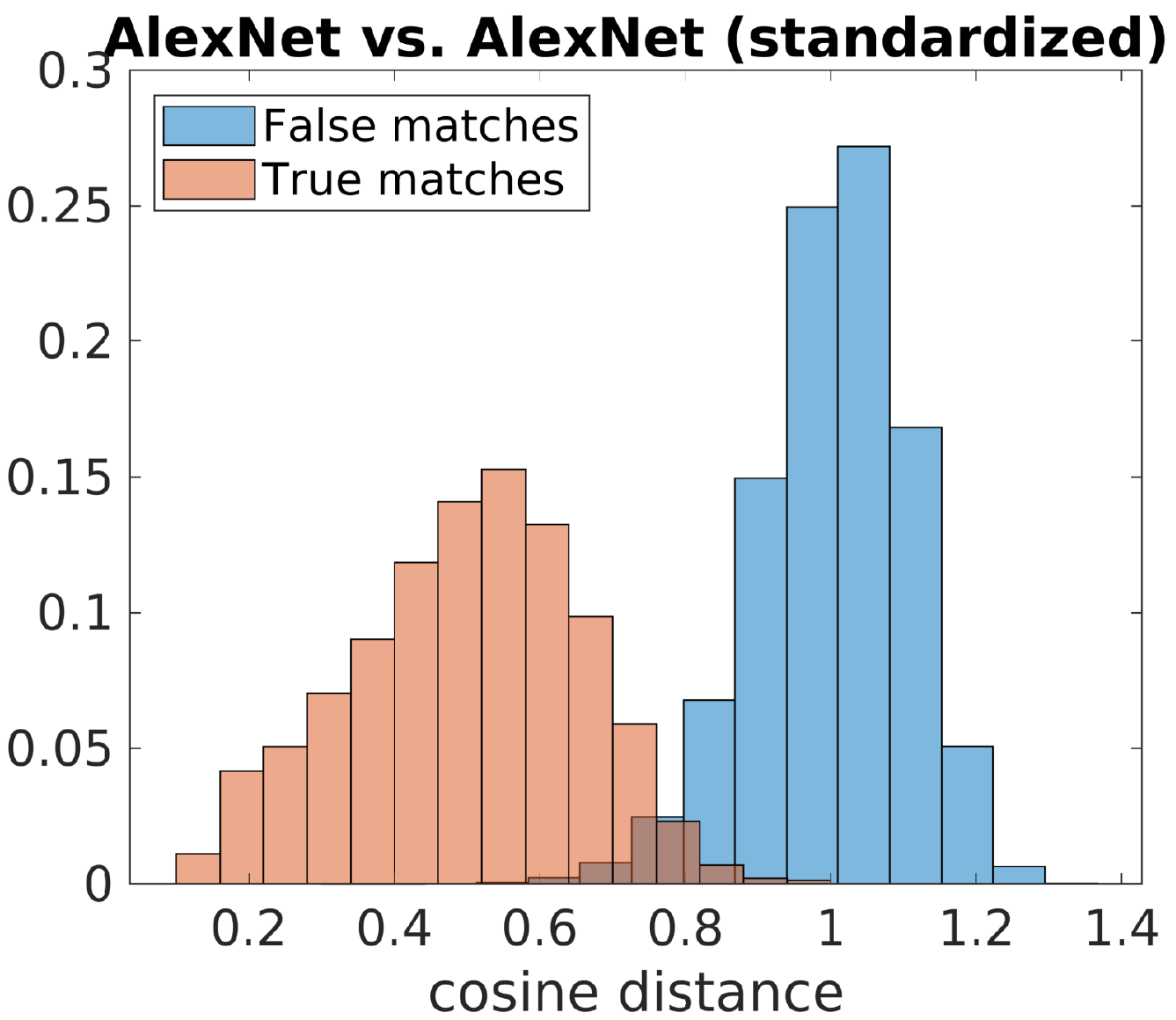} 
 \includegraphics[width=0.324\linewidth]{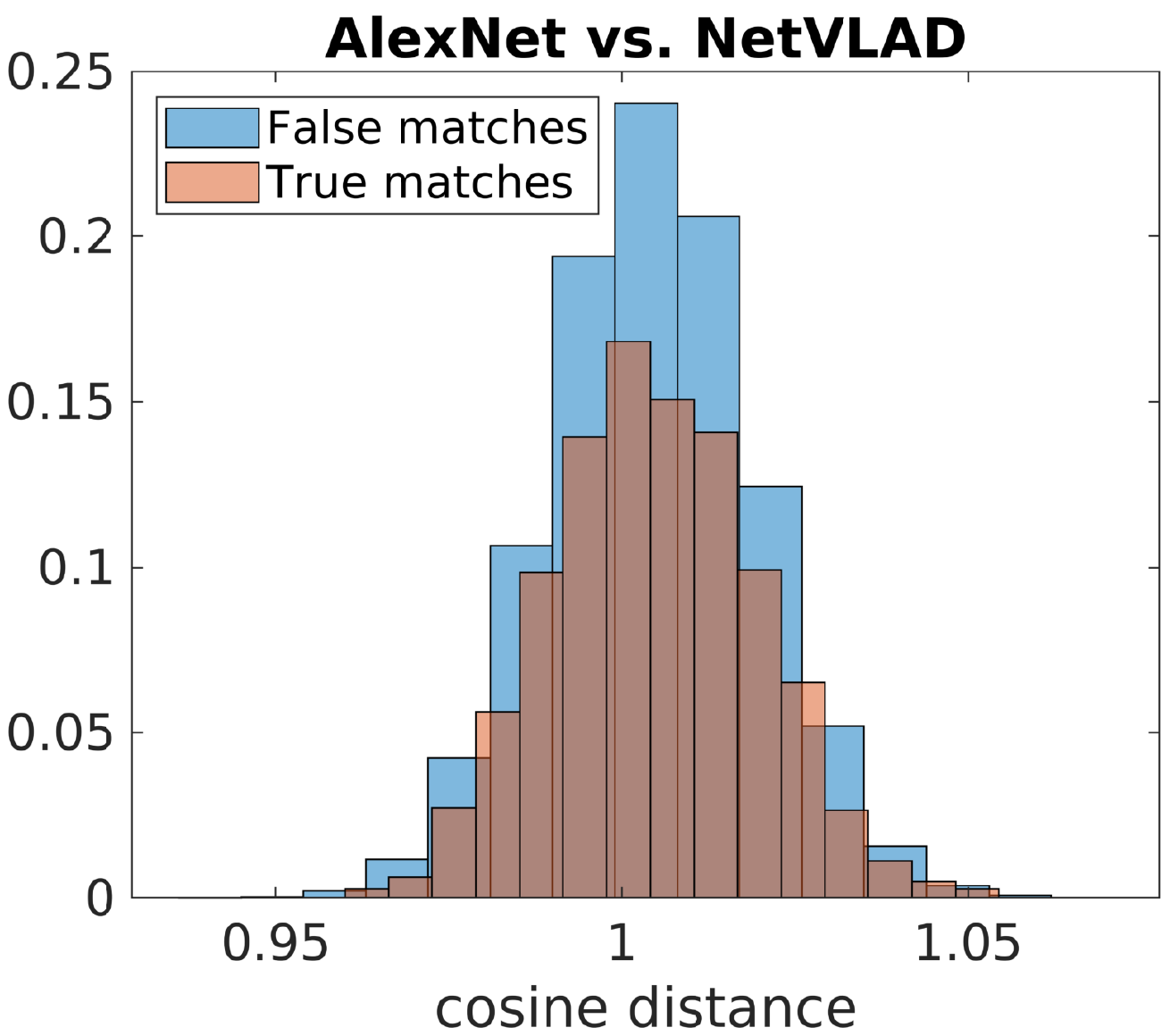}
 \caption{Distributions of distances for descriptors of the same place (true matchings) or of different places (false matchings).}
 \label{fig:descr_distr}
 \label{fig:global}
\end{figure}

% ============================================================================
\subsubsection{Vector space and random vectors for symbols}
For the vector space $\mathbb{V}$ we use real-valued d-dimensional vectors with $d$ in the order of thousands (we will use $d=4,096$ in our experiments).
However, based on the mechanisms how vectors are created and processed, most (if not all) vector entries will be in the range $[-1,1]$.
For measuring the similarity of vectors, we use cosine similarity (normalized dot-product).
Vectors can be created by three mechanisms:
(1)~Systematic encoding of a mathematical entity, a sensor measurement, or similar. Image encodings are topic of Sec.~\ref{sec:image_descriptors}, position encodings of Sec.~\ref{sec:pose_encoding}.
(2)~The VSA operations binding or bundling combine multiple vectors of space $\mathbb{V}$ to a new vector from the same space
(3)~Random vector are used to encode symbolic entities, e.g. to represent elements of a finite (enumerable) set of classes.
We create random vectors by sampling each dimension independently from the two-elemental set $\{-1,1\}$ with equal probability for each of the two values.
The reason for this very special initialization are the properties of the binding operator.

% ============================================================================
\subsubsection{Binding $\otimes$}
\label{sec:binding}
Binding $\otimes: \mathbb{V} \times \mathbb{V} \rightarrow \mathbb{V}$ is the first of two implemented HDC operations.
In the general context of HDC \cite{Kanerva09}, this operation is used to bind fillers to roles, e.g. to assign a particular value to a variable. 
More specifically, a vector $\in\mathbb{V}$ that represents the value is bound to another vector $\in\mathbb{V}$ that represents the variable.
(Vector representations for variables are also an example for the above mentioned symbolic entities that are encoded with random vectors.)
The result of binding is a new vector $\in\mathbb{V}$ with two important properties:

(1) It is \textit{not similar} to the two input vectors but allows to (approximately) \textit{recover} any of the input vectors given the output vector and the other input vector. 
In the general HDC context, recovering is done by an unbinding operator, for those HDC implementations where vectors are also (approximately) self inverse, unbinding and binding are the same operation (e.g. \cite{kanerva97,Gayler98}).
Self-inverse means: 
$\hspace{1cm} \forall x \in \mathbb{V}: x \otimes x = \mathbf{1}$
where $\mathbf{1}$ is the neutral element of binding in the space $\mathbb{V}$. 

(2) Binding is \textit{similarity preserving}. The distance of the output of binding a vector to two different vectors depends on the similarity of these vectors: \mbox{$\forall x,a,b \in \mathbb{V}: dist( x \otimes a, x \otimes b) \approx dist(a,b)$}

We use elementwise multiplication for binding two vectors.
In a vector space $\mathbb{V} = \{-1,1\}^d$, binding by element-wise multiplication is exactly self-inverse ($-1 \cdot -1 = 1 \cdot 1 = 1$) and the resulting vector has a large distance to each of the input vectors (since the sign is switched for each $-1$ entry in the other vector, which is expected to happen for about 50\% of the dimensions).
When using the vector space $[-1,1]^d$ instead of $\{-1,1\}^d$, both properties only hold approximately \cite{Schlegel2020}. 
However, this modification is required to implement the bundling operator.

% ============================================================================
\subsubsection{Bundling $\oplus$}
\label{sec:bundling}

The purpose of the bundling operation $\oplus: \mathbb{V} \times \mathbb{V} \rightarrow \mathbb{V}$ is to combine (``superpose'') two input vectors such that the output vector is \textit{similar} to both inputs. 
In almost all HDC implementations, the bundling operator is some kind of elementwise sum or averaging of the vector elements.

We use the elementwise sum.
When adding two vectors $z=x+y$, the vector $y$ can be seen as noise that disturbs the similarity of $x$ and $z$. 
In high-dimensional vectors spaces, random vectors are very likely almost orthogonal (a property called quasi-orthogonality) \cite{Neubert19}.
This has two important effects: 
(1)~If $x$ and $y$ are quasi-orthogonal and of similar magnitude, the influence of adding the noise $y$ to $x$ on the direction of $x$ is limited, in particular, $y$ does not point anywhere close to the opposite direction of $x$. This limits the influence of $y$ on the angular similarity between $z$ and $x$.
(2)~If the expected angle between random (unrelated) vectors is close to 90$^o$, than any considerably smaller angular distance indicates a relation between the compared vectors. 
Thus the noise $y$ can reduce the similarity but the remaining similarity will remain considerably about chance and $z$ and $x$ can be considered similar.

Bundling several image descriptors is the simplest way to use HDC for feature aggregation (cf. Sec.~\ref{sec:approach_global}). 
However, for more sophisticated encoding of information, we will combine the bundling and binding operators.
According to Kanerva \cite{kanerva14} bundling and binding should ``form an algebraic field or approximate a field''.
In particular, our bundling and binding are associative and commutative and binding distributes over bundling.

\subsubsection{Preprocessing of image descriptors}
\label{sec:image_descriptors}
We will use different image descriptors (e.g. NetVLAD \cite{netvlad} or DELF \cite{delf}) in our experiments. 
Each outputs a high-dimensional vector. We use Gaussian random projection to control the number of dimensions and to distribute information across dimensions. 
We use L2 normalization to standardize the descriptor magnitudes, followed by mean-centering. 

Very much in line with our requirements, typical image descriptors aim to encode multiple images of the same scene with similar vectors and those of different scenes with different  vectors.
A typical distribution can be seen in Fig.~\ref{fig:global} (left).
The middle part of this figure also illustrates the effect of mean-centering the descriptors (i.e. subtracting the mean descriptor of the database and query sets). This is known to considerably improve place recognition results (i.e. it improves the separation of the blue and the red distributions) \cite{SchubertNP20}. However, it also improves the quasi-orthogonality property for descriptors of different places.
We include mean-centering for both reasons. 

\begin{figure}
  \centering
 \includegraphics[width=1\linewidth]{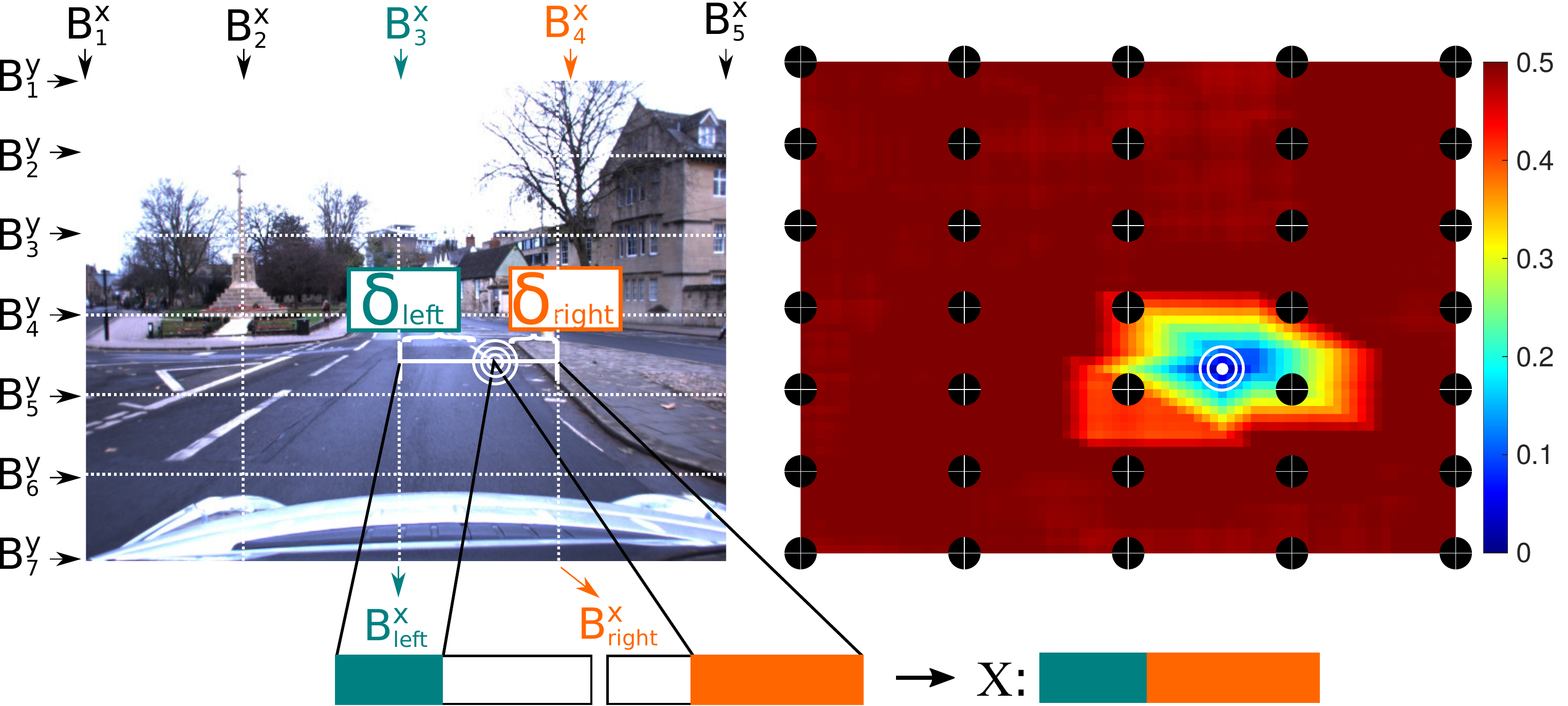}
 \caption{ Illustration of the pose encoding from Sec.~\ref{sec:pose_encoding}. \textit{(left)} Layout of basis vectors $B$ and combination of two basis vectors $B_{left}$ and $B_{right}$ to create encoding $X$ of the horizontal image position of the white marked location. 
 \textit{(right)} 
 The Hamming distance of this $\{-1,1\}^d$ vector to all other image location encodings (however, in the HDC framework, angular distance will be used). 
 }
 \label{fig:pose_encoding}
\end{figure}

\subsubsection{Encoding of positions}
\label{sec:pose_encoding}
The goal is to reflect the spatial distance of different landmarks with positions $(x,y)$ in the image by the angular distance of the pose vector encodings $\in\mathbb{V}$.
There are several alternatives for the creation of such encodings, this section will propose a simple yet flexible approach.

We create vectors $X$ and $Y$ to encode $x$ and $y$ independently and compute the final pose encoding by $P = X \otimes Y$.
This paragraph and Fig.~\ref{fig:pose_encoding} explain the creation of $X$. $Y$ is computed accordingly. To encode $x$ from range $[1,w]$ for an image of width $w$, we equally divide this range in $n_x$ subintervals and associate each border of subintervals with one of $n_x+1$ random basis vectors $B^X \in \{-1, 1\}^d$ (including the beginning of the first and the end of the last subinterval).
The encoding $X$ of $x$ is then computed by concatenating parts of the basis vectors $B^X_{left}$ and $B^X_{right}$ of the subinterval in which $x$ is located. With Matlab-style syntax this is:
\begin{equation}
\label{eq:poseX}
X = [ B^X_{left}(1:\alpha), B^X_{right}(\alpha+1:end) ]
\end{equation}
$\alpha$ is the splitting index based on the distances $\delta_{left}$,$\delta_{right}$ of $x$ to the two subinterval borders and the dimensionality $d$: 

\begin{equation}
\alpha =  \left \lfloor d \cdot \frac{\delta_{right}}{\delta_{left} + \delta_{right}} \right \rceil 
\end{equation}

This approach is flexible since parameters $n_x$ and $n_y$ can be used to weight the spatial distances in each direction for a particular application. In our place recognition experiments, we will use $n_x=4$ and $n_y=6$ which allows larger horizontal displacements of landmarks than vertical displacements. 
An example resulting distance map can be seen in Fig.~\ref{fig:pose_encoding}.
The normalized dot product of this encoding approximates the rectilinear distance (1-norm) of the encoded image locations.
The distortions in Fig.~\ref{fig:pose_encoding} result from interference between X and Y encodings.
It is important to note that although we divide the image in a grid, this approach is able to evaluate similarities \textit{across} the grid borders.

% ============================================================================
% ============================================================================
\subsection{HDC for feature aggregation}
\label{sec:feat_aggr}

% ============================================================================
\subsubsection{Unordered aggregation of multiple descriptors}
\label{sec:approach_global}
Based on the effect of bundling on similarities explained in Sec.~\ref{sec:bundling}, this operator can be used to combine the information of multiple image descriptors in a single vector $H$.
This can, e.g., be used to aggregate \textit{multiple} holistic image descriptors $H_i$, $i=1...k$ (e.g., NetVLAD, AlexNet-conv3, and DenseVLAD) for a single image. 
Fig.~\ref{fig:global} shows a typical distribution of similarities when comparing \textit{different} descriptors. Not surprisingly, computing the distance between a NetVLAD descriptor and an AlexNet-conv3 descriptor is not useful since they behave to each other very much like random vectors. 
However, as described in Sec.~\ref{sec:bundling}, this allows to safely bundle these different descriptors in a single vector (after the preprocessing from Sec.~\ref{sec:image_descriptors}):

\begin{equation}
\label{eq:global}
H=\bigoplus_{i=1}^k H_i = \sum_{i=1}^k H_i
\end{equation}

Basically, this is a simple averaging of descriptors. However, due to the quasi-orthogonality of different descriptor classes, evaluating the angular distance of such bundled vectors of two images by normalized dot product will approximate the average distance of the bundled descriptor classes if they were evaluated individually. 

We want to emphasize, that we do not claim a performance benefit of this approach compared to, e.g., concatenating dimensionality-reduced input descriptors (the experiments will show roughly equal performance). 
It is rather intended to illustrate the type of computation in the proposed HDC framework and its flexibility.
For example, if we want to be able to recover an approximate version of each input descriptor $H_i$ from the resulting combined descriptor, we can bind each input descriptor (before bundling) to a \textit{fixed} random vector $T_i$ that represents the descriptor type:
$H' = \bigoplus_{i=1}^k T_i \otimes H_i$. 
A particular descriptor of type $T_i$ can then be approximately recovered by unbinding: $H_i = T_i \otimes R$.

% ============================================================================
\subsubsection{Systematic encoding of local feature descriptors and positions}
\label{sec:approach_local}

This section describes an approach to compute a holistic descriptor $L \in \mathbb{R}^d$ that encodes a variably sized set of local descriptors $L_i \in \mathbb{R}^d$ together with their poses $(x_i,y_i)$, $i=1...k$.
$L$ and $L_i$ are from the same vector space (i.e. they have the same number of dimensions) and the angular distance of two holistic vectors $L^A$ and $L^B$ of images A and B will approximate the distance from an exhaustive pairwise comparison of the local features and their poses.

To generate a holistic descriptor $L$, all local feature descriptors are first preprocessed according to Sec.~\ref{sec:image_descriptors}. The mean-centering is done using all descriptors from the current image. 
Each pose $(x_i,y_i)$ is encoded in a vector $P_i \in \{-1,1\}^d$ as described in Sec.~\ref{sec:pose_encoding}.
The holistic descriptor of local features with poses is then computed by:
\begin{equation}
\label{eq:local}
 L = \bigoplus_{i=1}^k L_i \otimes P_i 
\end{equation}
To use this for image matching applications like place recognition, we compute a holistic descriptor for each image from all local features. The number and spatial arrangement of local features can vary between images. Holistic descriptors are then compared by normalized dot product. 

The result is an approximation of the exhaustive pairwise comparison of all feature pairs using their descriptor and spatial distances.
We want to give an intuition to this approximation using the example of comparing an image $A$ with descriptor \mbox{$L^A = \bigoplus_{i=1}^{k^A} L^A_i \otimes P^A_i$} and an image $B$ with descriptor \mbox{$L^B = \bigoplus_{i=1}^{k^B} L^B_i \otimes P^B_i$}.
From an HDC perspective, comparing $L^A$ and $L^B$ multiplies out and creates individual comparisons of the form
$L^A_i \otimes P^A_i$ vs. $L^B_j \otimes P^B_j$ and the overall similarity is the accumulated similarity of all such comparisons.
We can assume that each pose vector is quasi-orthogonal to each descriptor vector.
Based on this and the two properties of binding from Sec.~\ref{sec:binding}, the two vectors in each term can only be similar if the descriptors are similar ($L^A_i \approx L^B_j$) \textit{and} the poses are similar ($P^A_i \approx P^B_j$).
The normalized dot product of the holistic descriptors from Eq.~\ref{eq:local} is only an approximation of the exhaustive comparison since, in reality, multiplying-out is prevented by the required normalization of the vectors.
The later experiments in Fig.~\ref{fig:local_bundling_nordland} will show the effect on local feature similarity. For more information, please refer to the HDC literature, e.g. \cite{Kanerva09,Schlegel2020}. 

% ============================================================================
\subsubsection{Extensions and framework character}
\label{sec:approach_extension}

The concept of bundling and binding to aggregate information can be easily extended to other information than position of local features.
For example, we can integrate information about local feature scale or orientation by binding to appropriate encodings (e.g., similarly created as X in eq.~\ref{eq:poseX}).
This also applies to sequences of images. Exploiting the similarity of temporally neighbored images can significantly improve place recognition performance in mobile robotics  \cite{seqslam,Hansen2014,NeubertSP19}. SeqSLAM \cite{seqslam} is a simple yet powerful approach that accumulates similarities of short sequence of image comparisons.
This requires the computation of all similarities individually. An appropriate bundling of image descriptors (each bound to its position within the sequence) can achieve very similar results with a \textit{single} vector comparison \cite{Neubert19, Schlegel2020}.
We want to emphasize that our choices of $\mathbb{V}$, bundling, and binding are only one possible HDC implementation, there are several others available \cite{Schlegel2020} that can potentially also be used in the presented approaches.

%%%%%%%%% BODY TEXT
\section{Experimental Results}
\label{sec:experiments}

\subsection{Experimental setup}

We will evaluate the HDC approach on standard place recognition datasets from mobile robotics. 
We use 23 sequence comparisons from six datasets with different characteristics regarding environment, appearance changes, single or multiple visits of places, possible stops, or viewpoint changes: \textbf{Nordland1k}~\cite{ds_nordland}, \textbf{StLucia} (Various Times of the Day)~\cite{ds_stlucia}, \textbf{CMU} Visual Localization~\cite{ds_cmu}, \textbf{GardensPointWalking}\footnote{\url{https://goo.gl/tqmWyq}}, \textbf{OxfordRobotCar}~\cite{ds_robotcar}, and \textbf{SFUMountain}~\cite{sfumountain}.
For OxfordRobotCar, we sampled sequences at 1Hz with the recently published accurate ground truth data~\cite{rtk}.
For Nordland1k, we sampled 1k images of unique places from each season (without tunnels). 

We decided to use DELF \cite{delf} for local features since it provides good results using standard exhaustive pairwise comparison (see Table~\ref{tab:res_local}, in our (not shown) experiments it performed better than, e.g., \cite{d2net} and \cite{suenderhauf15a}). 
Moreover, with DELG, there is already a deep-learned holistic descriptor available for comparison that builds upon DELF.

We compare against the following descriptors.
NetVLAD \textbf{NV} \cite{netvlad}: We use the authors' VGG-16 version\footnote{\url{https://github.com/Relja/netvlad}} with whitening trained on the Pitts30k dataset (4,096-D). 
AlexNet \textbf{AN} \cite{alexnet}: We use the conv3 output of Matlab’s ImageNet model and the full 65k dimensional descriptor. 
DenseVLAD \textbf{DV} \cite{Torii15}: We use the authors' version\footnote{\url{http://www.ok.ctrl.titech.ac.jp/~torii/project/247/}} with 128-dimensional SIFT descriptors and 128 words trained on 24/7 Tokyo dataset, as well as PCA projection to 4,096-D.
\textbf{DELG} \cite{CaoAS20}: We use the implementation from TensorFlow models with ResNet101 trained on a subset of the Google Landmarks Dataset v2 (GLDv2-clean) which was amongst best in \cite{CaoAS20}.

Besides these holistic descriptors from the literature, we use two ways to exhaustively compute image similarities from local features:
\textbf{DELF} \cite{delf}: We use the implementation from TensorFlow Hub\footnote{\url{https://tfhub.dev/google/delf/1}}. For each image we extract the 200 1024-dimensional descriptors with highest score at scale 1. The descriptors are standardized per image \cite{SchubertNP20}. Following \cite{suenderhauf15a}  image similarity $s$ is computed from mutual matchings $M$ for exhaustive pairwise comparison of $n^{DB}, n^Q$ features with uniform position weighting $p_{ij}=1$: 
 \begin{align}
    s = \frac{1}{\sqrt{n^{DB} \cdot n^Q}} \sum _{i,j\in M} p_{ij} \cdot sim(L^{DB}_i, L^Q_j)
 \end{align}
\textbf{DELF-pos}: Same as before but we incorporate spatial distance of local features with weighting $p_{ij}$ as follows

 \begin{align}
     p_{ij} = \min(&\max(0, 1-\frac{|x_i-x_j|}{w/n_x}), \max(0, 1-\frac{|y_i-y_j|}{h/n_y}))
 \end{align}
$w,h$ are image dimensions, $n_y, n_x$ are the same as for HDC.

Further, we compare against four existing methods that create a holistic descriptor from local features. We combine all with DELF:
\textbf{DELF-V} is DELF with VLAD \cite{vlad}: We use the same 200 1024-D DELF descriptors as before in a VLAD representation that is trained from all datasets that are solely used as database as well as additional night and winter images from Oxford and the \"Orebro Seasons dataset~\cite{ds_seasons}. 
We use VLFeat for kmeans and VLAD implementations, vocabulary size $64$, VLAD with hard assignment, L2-normalized components, as well as L2-normalization of the $64 \cdot 1024=65536$-D descriptors. 
\textbf{DELF-V-PCA}: Same as above but with PCA projection to 4096-D, PCA is trained on the vocabulary training data. 
\textbf{DELF-MV}: Following \cite{ArandjelovicZ13}, we compute a MultiVLAD representation by concatenating 14 VLAD-PCA descriptors over different regions of the images as described in \cite{ArandjelovicZ13}.
\textbf{DELF-Grid}: Computes a regular grid of $102$ DELF descriptors at scale 1. The descriptors are dimensionality reduced with PCA to 40-D (provided by DELF implementation) and concatenated to get a $102 \cdot 40 = 4080$-D holistic descriptor.

For evaluation, we compute pairwise similarity matrices between database and query image sets and compare them to ground-truth knowledge about place matchings using a series of thresholds. We report average precision (AP) computed as area under the resulting precision-recall curve, as well as achieved recall using the best k-matchings.

\subsection{Evaluation of unordered aggregation of multiple holistic descriptors}
\label{sec:res_global}
\begin{figure}
 \includegraphics[width=\linewidth]{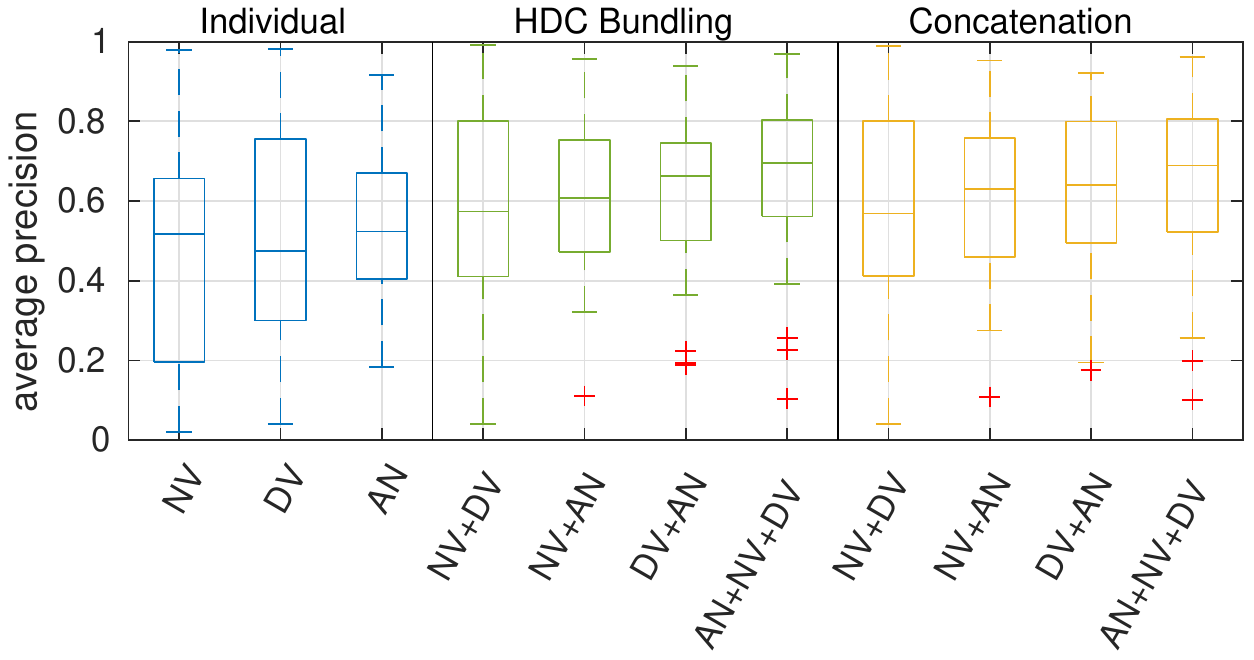}
 \caption{Average precision statistics over a large number of datasets. A simple HDC bundling of multiple descriptors can considerably improve the performance. However, this is also possible by concatenating descriptors.}
 \label{fig:global_boxplot}
\end{figure}

For a first impression of the potential of the simple HDC operators, the boxplots in Fig.~\ref{fig:global_boxplot} show a considerable improvement in median average precision as well as outlier statistics, when combining multiple different holistic descriptors as described in Sec-~\ref{sec:approach_global}. 
Before bundling, all descriptors are projected to 4,096-D using Gaussian random projection.
The boxplots include all datasets from Table~\ref{tab:res_local} (details not shown).
However, very similar results can be achieved by concatenating the descriptors (right part of Fig.~\ref{fig:global_boxplot}).

\definecolor{notworking}{rgb}{.2,.2,.2} 
\definecolor{supergood}{rgb}{.2,.6,.2}
\definecolor{good}{rgb}{.5,.99,.5} 
\definecolor{equal}{rgb}{0.1,0.1,0.1}  
\definecolor{bad}{rgb}{.9,.7,.1} 
\definecolor{superbad}{rgb}{.9,.1,.1}    
\fboxrule0.65pt

\begin{table*}[tb]
\caption{Average precision of the proposed DELF-HDC and other methods at standard place recognition datasets from mobile robotics. The best result of all holistic descriptors per dataset is highlighted, DELF and DELF-Pos are exhaustive local comparisons.}
\centering 
\resizebox{1\textwidth}{!}{%
  \boldmath
  \begin{tabular}{|lll|| c c c c c c c c c| c c|}
  \hline
  \textbf{Dataset} & \textbf{DB} & \textbf{Query} & \textbf{NV} & \textbf{AN} & \textbf{DV} & \textbf{DELG} & \textbf{DELF-V} & \textbf{DELF-V-PCA} & \textbf{DELF-MV} & \textbf{DELF-Grid} & \textbf{DELF-HDC} & \textbf{DELF} & \textbf{DELF-Pos} \\ 
  &  &  &  &  &  &  &  &  &  & & (ours) & (exhaustive) &  (exhaustive) \\ 
\hline
GardensPointWalking & day$\_$right & day$\_$left &  \textbf{0.99}  & 0.46  & 0.98  & 0.95  & 0.93  & 0.95  & 0.94  & 0.56  & 0.82  & 0.94  & 0.95  \\ 
 & day$\_$right & night$\_$right & 0.59  & 0.62  & 0.52  & 0.44  & 0.41  & 0.41  & 0.62  & 0.70  &  \textbf{0.79}  & 0.52  & 0.80  \\ 
 & day$\_$left & night$\_$right &  \textbf{0.48}  & 0.12  & 0.22  & 0.32  & 0.24  & 0.26  & 0.38  & 0.23  & 0.46  & 0.29  & 0.73  \\ 
\hline
OxfordRobotCar & 2014-12-09-13-21-02 & 2015-05-19-14-06-38 & 0.89  & 0.77  & 0.85  & 0.70  & 0.75  & 0.76  & 0.87  & 0.87  &  \textbf{0.90}  & 0.94  & 0.97  \\ 
 & 2014-12-09-13-21-02 & 2015-08-28-09-50-22 & 0.66  & 0.41  & 0.62  & 0.23  & 0.33  & 0.33  & 0.36  & 0.56  &  \textbf{0.70}  & 0.37  & 0.60  \\ 
 & 2014-12-09-13-21-02 & 2014-11-25-09-18-32 &  \textbf{0.91}  & 0.67  & 0.90  & 0.68  & 0.77  & 0.78  & 0.81  & 0.88  & 0.80  & 0.79  & 0.88  \\ 
 & 2014-12-09-13-21-02 & 2014-12-16-18-44-24 & 0.11  & 0.27  & 0.11  & 0.12  & 0.05  & 0.06  & 0.09  & 0.66  &  \textbf{0.78}  & 0.17  & 0.64  \\ 
 & 2015-05-19-14-06-38 & 2015-02-03-08-45-10 &  \textbf{0.93}  & 0.84  & 0.33  & 0.72  & 0.44  & 0.38  & 0.71  & 0.66  & 0.75  & 0.81  & 0.92  \\ 
 & 2015-08-28-09-50-22 & 2014-11-25-09-18-32 & 0.59  & 0.34  & 0.46  & 0.38  & 0.43  & 0.38  & 0.52  & 0.57  &  \textbf{0.69}  & 0.53  & 0.72  \\ 
\hline
SFUMountain & dry & dusk & 0.48  & 0.54  &  \textbf{0.79}  & 0.34  & 0.30  & 0.29  & 0.45  & 0.76  &  \textbf{0.79}  & 0.74  & 0.81  \\ 
 & dry & jan & 0.22  & 0.40  & 0.63  & 0.10  & 0.11  & 0.10  & 0.24  &  \textbf{0.64}  & 0.55  & 0.61  & 0.82  \\ 
 & dry & wet & 0.40  & 0.42  &  \textbf{0.75}  & 0.25  & 0.22  & 0.21  & 0.29  & 0.66  & 0.73  & 0.71  & 0.82  \\ 
\hline
CMU & 20110421 & 20100901 & 0.71  & 0.52  & 0.69  &  \textbf{0.80}  & 0.56  & 0.61  & 0.69  & 0.59  & 0.74  & 0.80  & 0.82  \\ 
 & 20110421 & 20100915 & 0.77  & 0.65  & 0.76  &  \textbf{0.78}  & 0.59  & 0.62  & 0.73  & 0.70  & 0.73  & 0.75  & 0.77  \\ 
 & 20110421 & 20101221 & 0.54  & 0.36  & 0.49  & 0.59  & 0.54  & 0.57  & 0.55  & 0.49  &  \textbf{0.63}  & 0.62  & 0.67  \\ 
 & 20110421 & 20110202 & 0.62  & 0.39  & 0.49  & 0.64  & 0.42  & 0.44  & 0.51  & 0.45  &  \textbf{0.71}  & 0.72  & 0.82  \\ 
\hline
Nordland1k & spring & winter & 0.02  & 0.25  & 0.06  & 0.07  & 0.03  & 0.03  & 0.05  & 0.08  &  \textbf{0.74}  & 0.54  & 0.86  \\ 
 & spring & summer & 0.20  & 0.66  & 0.43  & 0.45  & 0.29  & 0.27  & 0.28  & 0.67  &  \textbf{0.72}  & 0.45  & 0.71  \\ 
 & summer & winter & 0.05  &  \textbf{0.57}  & 0.05  & 0.16  & 0.06  & 0.05  & 0.04  & 0.21  & 0.46  & 0.17  & 0.44  \\ 
 & summer & fall & 0.53  &  \textbf{0.92}  & 0.82  & 0.79  & 0.54  & 0.49  & 0.51  & 0.87  & 0.89  & 0.83  & 0.91  \\ 
\hline
StLucia & 100909$\_$0845 & 180809$\_$1545 & 0.08  &  \textbf{0.46}  & 0.28  & 0.15  & 0.10  & 0.10  & 0.25  & 0.43  & 0.44  & 0.29  & 0.45  \\ 
 & 100909$\_$1000 & 190809$\_$1410 & 0.19  & 0.57  & 0.46  & 0.22  & 0.37  & 0.39  & 0.61  & 0.55  &  \textbf{0.63}  & 0.48  & 0.64  \\ 
 & 100909$\_$1210 & 210809$\_$1210 & 0.61  & 0.66  &  \textbf{0.84}  & 0.63  & 0.76  & 0.77  & 0.81  & 0.62  & 0.69  & 0.65  & 0.68  \\ 
\hline
worst case &  &  & 0.02  & 0.12  & 0.05  & 0.07  & 0.03  & 0.03  & 0.04  & 0.08  &  \textbf{0.44}  & 0.17  & 0.44  \\ 
best case &  &  &  \textbf{0.99}  & 0.92  & 0.98  & 0.95  & 0.93  & 0.95  & 0.94  & 0.88  & 0.90  & 0.94  & 0.97  \\ 
average case (mAP) &  &  & 0.50  & 0.52  & 0.55  & 0.46  & 0.40  & 0.40  & 0.49  & 0.58  &  \textbf{0.70}  & 0.60  & 0.76  \\ 

  \hline
  \end{tabular}}
  \label{tab:res_local}
\end{table*}

\begin{figure}
 \includegraphics[width=\linewidth]{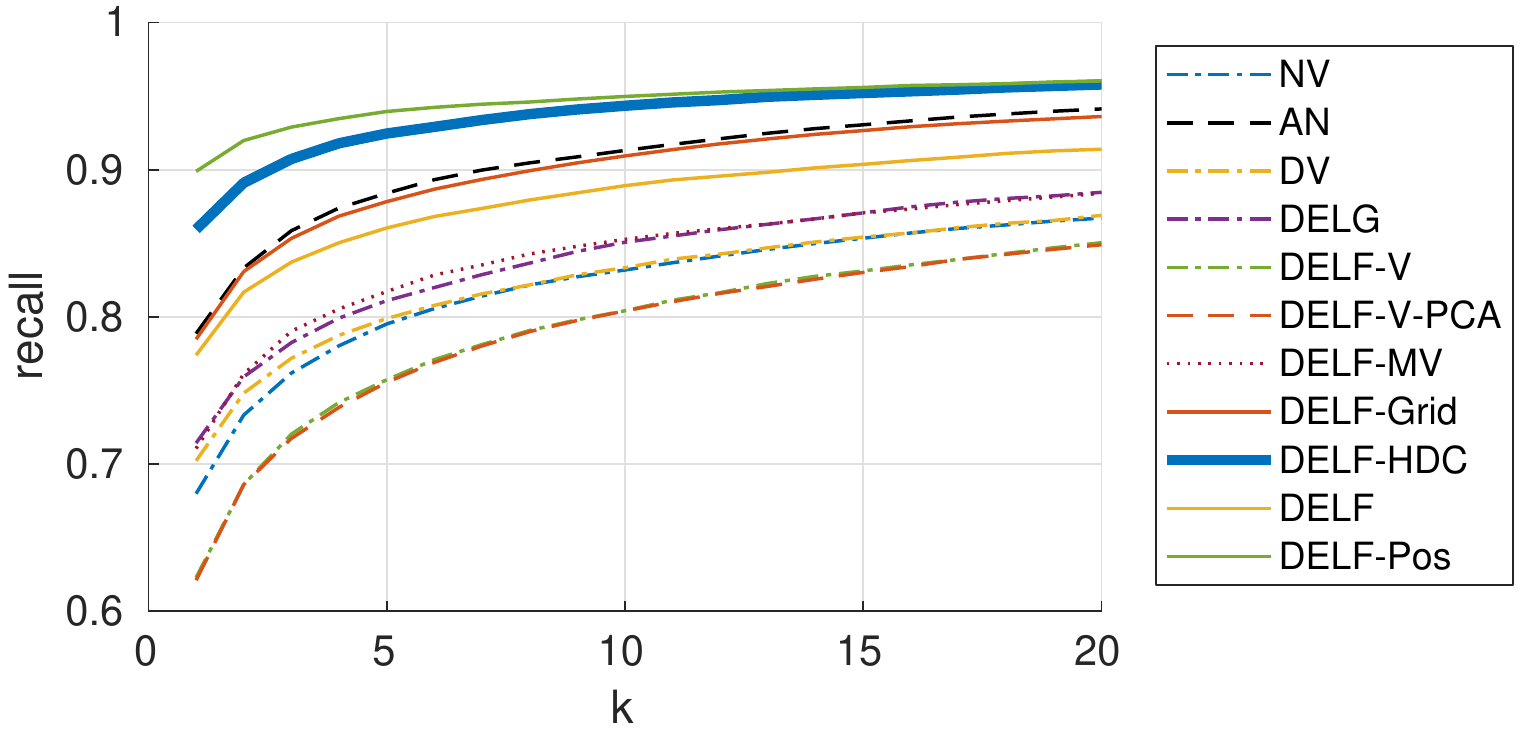}
 \caption{Achieved recall using the best k matchings. Averaged over all datasets from Table~\ref{tab:res_local}. DELF and DELF-Pos are exhaustive local comparisons, all others are fast holistic descriptor.}
 \vspace{-0.5cm}
 \label{fig:local_recall}
\end{figure}

\subsection{Evaluation of systematic aggregation of local descriptors and positions}

We implement the HDC approach to local feature aggregation from Sec.~\ref{sec:feat_aggr} using the same 200 highest-scored DELF descriptors as for the compared methods.
We refer to our approach as \textbf{DELF-HDC}. We use $4,096$ dimensional vectors and set spatial weighting parameters $n_x=4$ and $n_y=6$ to allow more horizontal than vertical viewpoint deviation (in accordance with the mobile robot place recognition task).
An evaluation of all three parameters follows in Sec.~\ref{sec:res_paramters}.

\subsubsection{Place recognition performance}
Table~\ref{tab:res_local} shows the average precision for each of the compared methods on each dataset. The proposed HDC-DELF approach is the best holistic descriptor for 11 of 23 comparisons and provides the best average case (+20\% to runner-up) and worst case performance (3.6x of runner-up) (this excludes DELF and DELF-Pos which are exhaustive local comparisons and much more time consuming, cf. Sec.~\ref{sec:runtime}).
Fig.~\ref{fig:local_recall} evaluates the application for candidate selection for visual pose estimation \cite{Sattler2018}. Again, DELF-HDC is only outperformed by the (prohibitively expensive) exhaustive pairwise comparison of DELF-Pos features.

\subsubsection{The importance of binding}
\begin{figure}[t]
 \centering
 \includegraphics[width=0.8\linewidth]{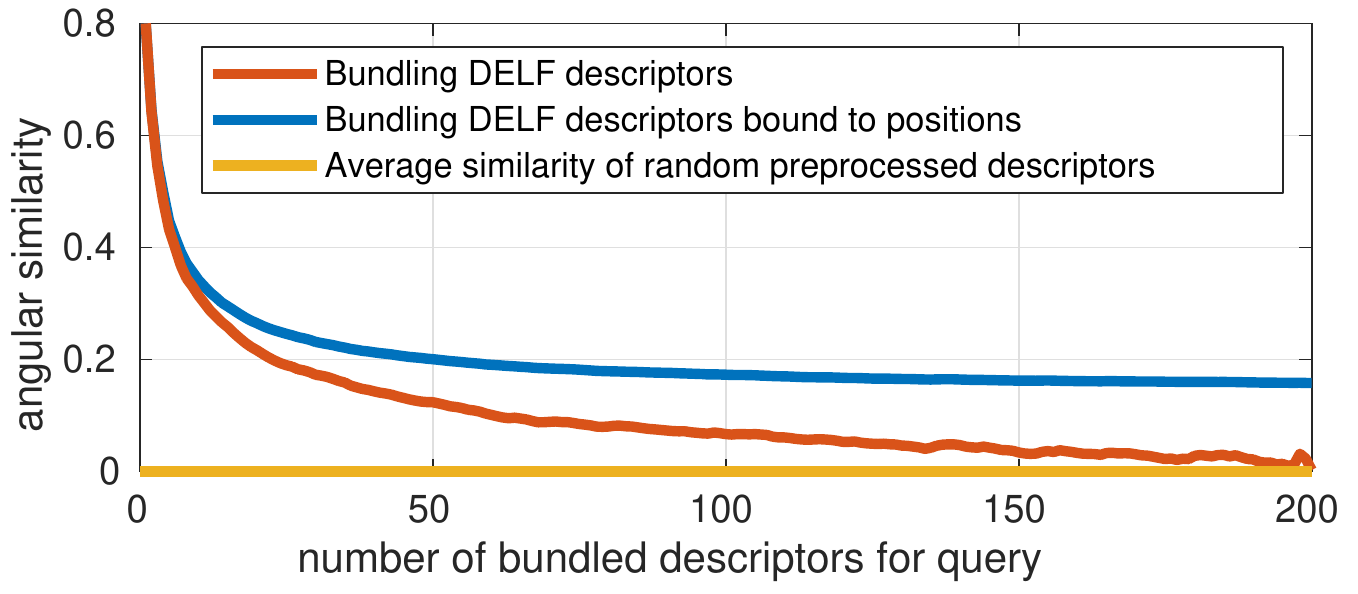}
 \caption{The similarity to an included single true matching in a bundle with an increasing number of distracting descriptors stays considerably above the average similarity of descriptors.}
 \label{fig:local_bundling_nordland}
\end{figure}

Why does this simple HDC approach work? 
To provide some intuition, Fig.~\ref{fig:local_bundling_nordland} shows the outcome of a simplified image matching experiment:
We take the encoding of a \textit{single} local feature $f_{DB}$ in a database image and find the most similar feature $f'_Q$ in a truly matching query image.
The similarity of their HDC encodings is shown at x-axis=1, for all other points on the curve, we bundle $f'_Q$ with an increasing number of other feature encodings from the query image, that 
act as noise on the similarity of the only true matching $f_{DB}$ and $f'_Q$.
Fig.~\ref{fig:local_bundling_nordland} shows average results on Nordland spring-summer for this experiment.
As illustrated by the yellow line, after preprocessing of descriptors, the expected or average similarity (normalized dot-product) of random pairs of descriptors from an image pair is close to 0, they are quasi-orthogonal.
A simple bundling of preprocessed descriptors (similar to the experiments in Sec.~\ref{sec:res_global}) is able to maintain a similarity considerably above chance for a few vectors (red).
When we additionally use binding to control similarities (according to the properties from Sec.~\ref{sec:bundling}) by incorporating additional position information, we can bundle significantly more ``noise'' vectors and still maintain a considerable similarity (compared to random pairs) of $f_{DB}$ and $f'_Q$ (blue).
Similar effects can be expected when binding with scale or sequential information.

\subsubsection{Properties and parameter evaluation}
\label{sec:res_paramters}
\begin{figure}
 \includegraphics[width=0.49\linewidth]{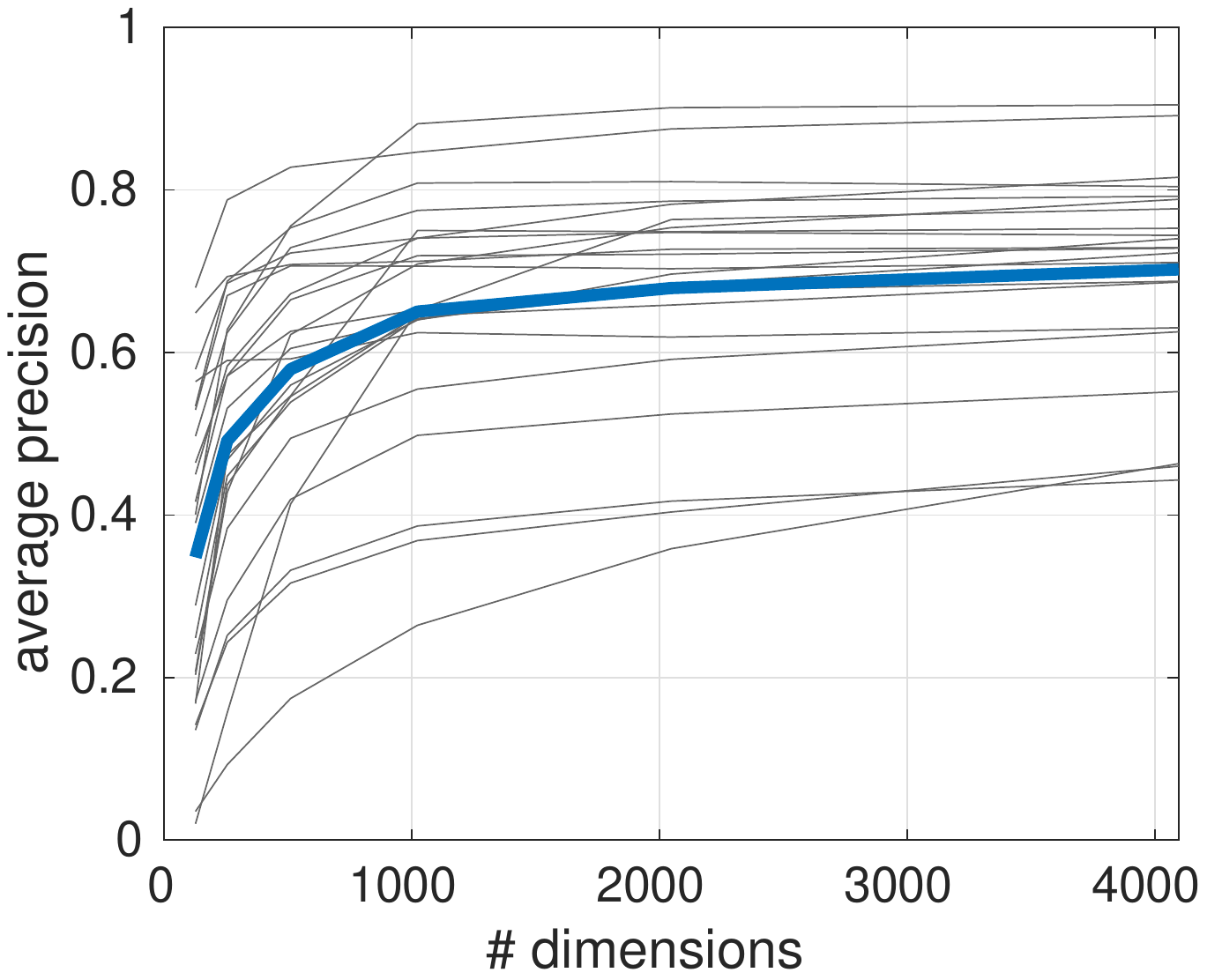}
 \includegraphics[width=0.49\linewidth]{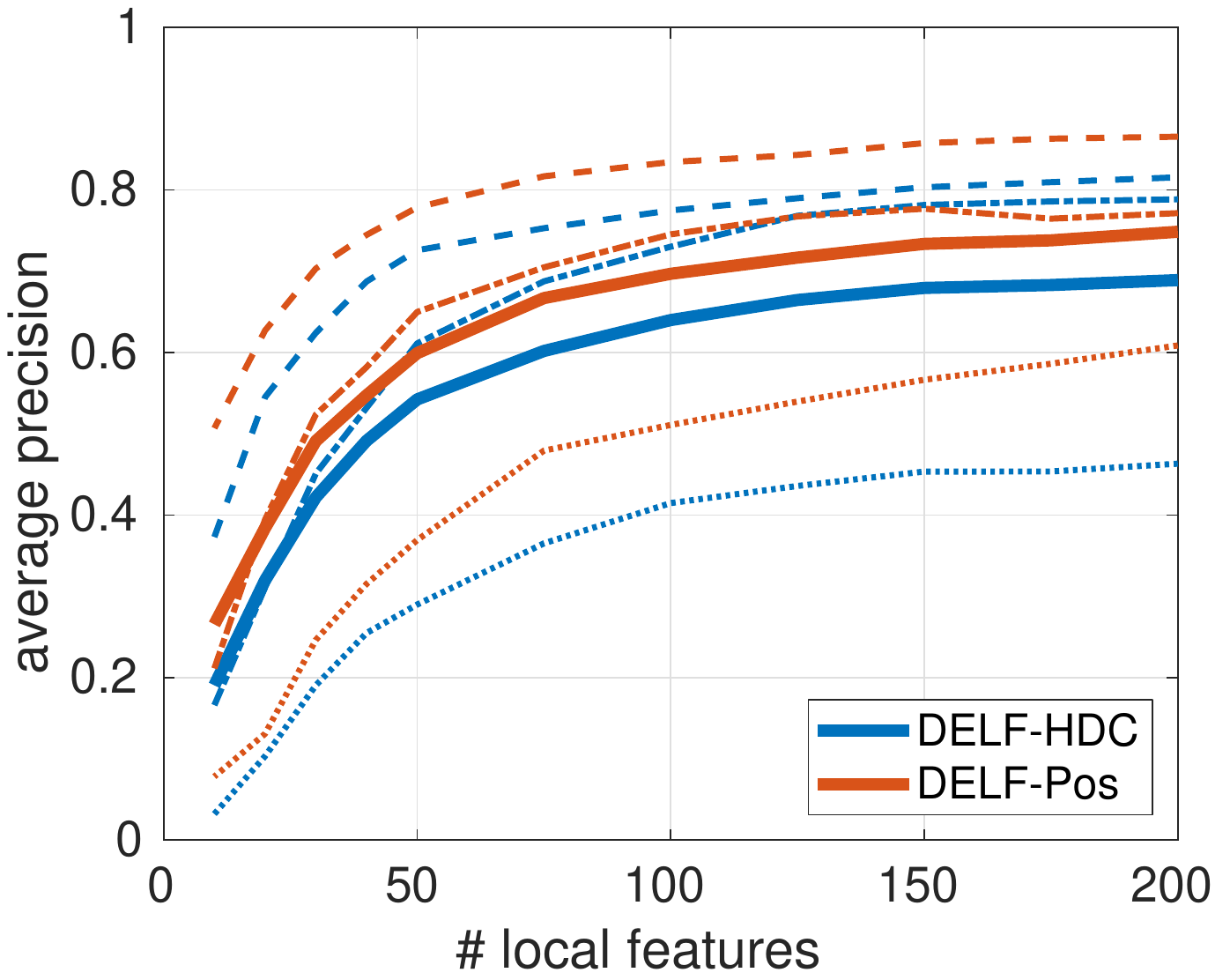}
 \caption{\textit{(left)} Average precision statistics over a varying number of dimensions in the HDC representations. Blue is the mean of all individual (gray) dataset evaluations (all datasets).
 \textit{(right)} Influence of a varying number of used features for the exhaustive and the HDC approach on GardensPointWalking, thin curves are individual comparisons, thick curves are means.
 }
 \label{fig:local_nDims}
 \label{fig:local_nFeats}
\end{figure}

A basic assumption in HDC is a high-dimensional vector space. The left part of Fig.~\ref{fig:local_nDims} evaluates the performance for a varying number of used dimensions. 
With $\ge$512 dimensions, the mean performance is equal or above the compared holistic algorithms. However, the lower the number of dimensions, the larger the variation on the individual datasets.

The right part of Fig.~\ref{fig:local_nFeats} evaluates the dependency on the number of features on the GardensPointWalking dataset. 
As expected, the performance of HDC-DELF increases with increasing number of features.
The roughly consistent distance to the exhaustive comparison (DELF-Pos) indicates that the capacity of the HDC representation is not exceeded.

\begin{figure}
 \includegraphics[width=1\linewidth]{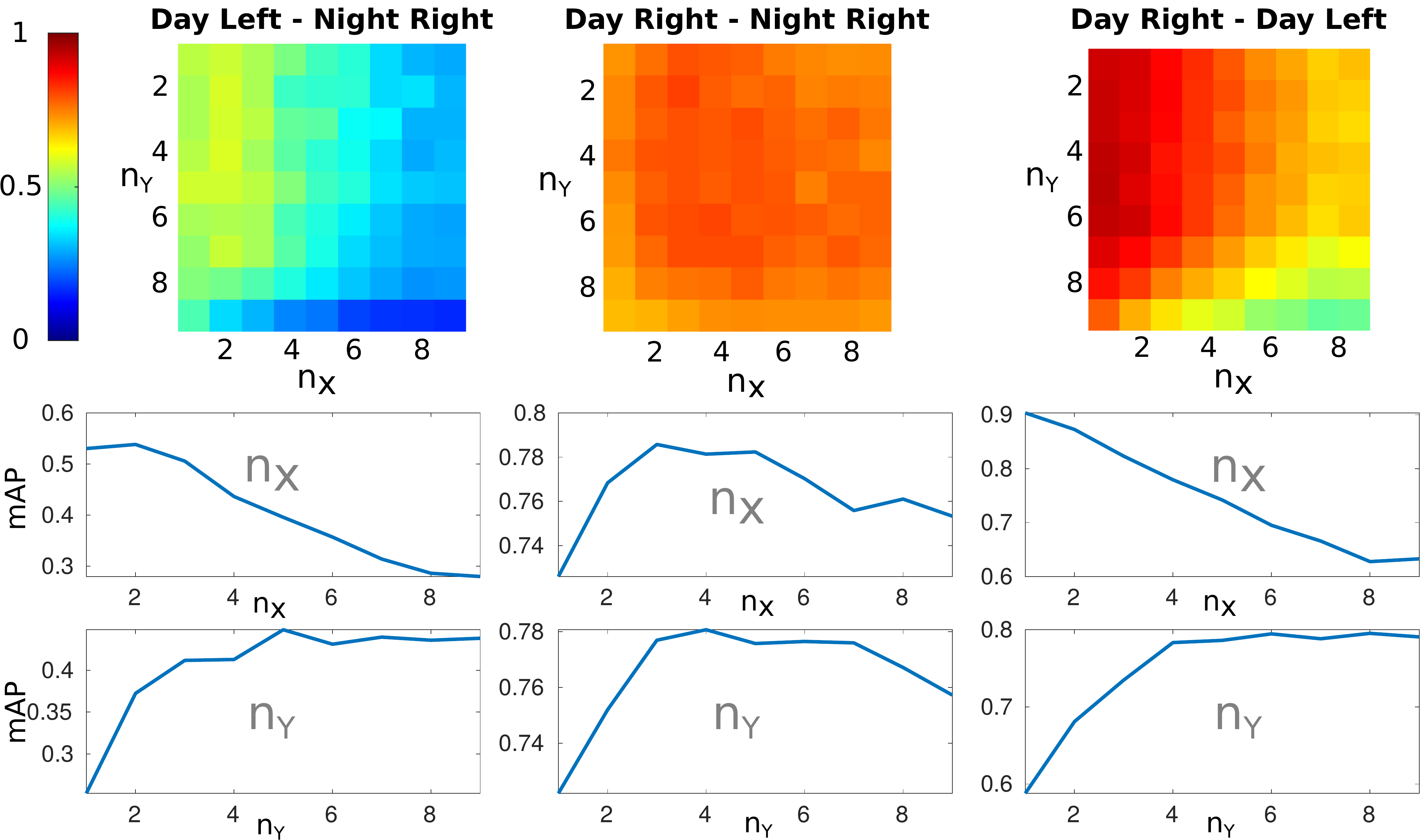}
 \caption{The heatmaps show average precision for different combinations of parameters $n_x$ and $n_y$ on GardensPointWalking. The graphs below show the mean column (for $n_x$) and row (for $n_y$) values of the heatmaps. To address viewpoint changes in the left-right combinations, a lower $n_x$ should be used.}
 \label{fig:local_nxny}
\end{figure}

In the presented HDC-DELF approach, parameters $n_x$ and $n_y$ can be used to control the sensitivity to viewpoint changes. 
Fig.~\ref{fig:local_nxny} shows the results for all combinations of values from range $\{1,2,...9\}$ for the three sequence comparisons of the GardensPointWalking dataset. 
This dataset is particularly interesting for this evaluation since it provides several sequences from day and night of a hand-held camera on the same pathway, but either on the left side or the right side of the pathway which results in a considerable horizontal viewpoint change.
For place recognition where only small viewpoint changes are expected, higher values (e.g. 7) of these parameters are preferable since they assign more different encodings to features  at large spatial distance (cf. Fig.~\ref{fig:pose_encoding}). To account for larger horizontal viewpoint changes, smaller values for $n_x$ can be use. 
We use the same parameters $\{n_y=6, n_x=4\}$ for all datasets in Table~\ref{tab:res_local}. However, tuning these parameters to a particular dataset is intuitive and can considerably improve results, e.g. with $n_x=2$ we can achieve AP=0.59 (+0.13) on the GardensPointWalking Day Left - Nigh Right comparison.

\subsection{Computational effort}
\label{sec:runtime}
Using holistic descriptors allows to compare two images by a single vector comparison with normalized dot-product.
This is significantly more efficient than exhaustive pairwise comparison of local features, e.g. in our setup with 200 local features, we would have to compute $200\cdot200=40,000$ vector distances per image comparison. Even if the local vectors are smaller and we can use ANN \cite{Li2020} techniques, there remains a discrepancy.

With the presented HDC approach, computing a holistic HDC descriptor from $n$ landmarks requires $n-1$ vector sums and $2n$ elementwise multiplications, one within the pose encoding and one for binding to the pose. The pose encoding also requires two times a concatenation of vectors.
Additionally, the descriptor preprocessing requires L2 normalization and mean-centering as well as potentially a projection to the used vector space. The latter might be the most time-consuming step in the overall computation.
In the HDC approach, binding and bundling operations can be computed by a single run over the vector. They operate locally on the vectors, i.e. only corresponding vector dimensions influence each other. This allows for massive parallelization. 
Since we work with distributed representations, an approximate similarity of holistic vectors can be easily computed by evaluating only a reduced number of dimensions (cf. Fig.~\ref{fig:local_nDims}).
We did not yet evaluate the combination with techniques like product quantization \cite{JegouDS11} for very large scale image retrieval (the largest used sequences were of size 4k).

\section{Conclusions}
We presented HDC as a simple to implement and flexible approach to combine descriptors and other information.
The presented HDC architecture was used to implement the HDC-DELF approach that combines local descriptors with their poses in a single holistic descriptor.
It can benefit from the advantages of local features like viewpoint and occlusion robustness, as well as feature selection (attention), without the drawbacks of exhaustive pairwise comparisons or fixed grid layouts. 
Parameters $n_x$,$n_y$ can be used to adjust the weighting of the pose similarity computation (which works \textit{across} grid borders).  
The HDC-DELF approach show improved performance on a series of standard place recognition datasets from mobile robotics on average and in worst case.

We consider HDC as a flexible framework since it allows to integrate further information like scale or orientation of local features, or to aggregate information across multiple images.
It can be combined with different existing and future image descriptors.
Further, the notation in HDC operations like bundling and binding allows to potentially improve our simple implementation using other HDC architectures (e.g., from \cite{Schlegel2020}).

%%%%%%%%% References
{\small
% \bibliographystyle{ieee_fullname}
% \bibliography{HDC}

}
\end{document}